# Detecting Image Forgeries using Geometric Cues

**Lin Wu,Yang Wang**

*Tianjin University, China*

**ABSTRACT**

This chapter presents a framework for detecting fake regions by using various methods including watermarking technique and blind approaches. In particular, we describe current categories on blind approaches which can be divided into five: pixel-based techniques, format-based techniques, camera-based techniques, physically-based techniques and geometric-based techniques. Then we take a second look on the geometric-based techniques and further categorize them in detail. In the following section, the state-of-the-art methods involved in the geometric technique are elaborated.

**INTRODUCTION**

Today's digital technology has begun to erode our trust on the integrity of the visual imagery since image editing software can generate highly photorealistic images (Farid, 2009). Doctored photographs are appearing with a growing frequency and sophistication in tabloid magazines, mainstream media outlets, political campaigns, photo hoaxes, evidences in a courtroom, insurance claims, and cases involving scientific fraud (Farid, 2009). With the rapid advancement in image editing software, photorealistic images will become increasingly easier to be generated and it becomes difficult for people to differentiate them from photographic images (Lyu & Farid, 2005). If we are to have any hope that photographs can hold the unique stature of being a definitive recording of events, we must develop technologies that can detect the tampered images. Therefore, authenticating the integrity of digital image's content has become particularly important when images are used as critical evidence in journalism and security surveillance applications.

Over the past several years, the field of digital forensics has emerged to authenticate digital images by enforcing several authentication methods. The presence or absence of the watermark in interpolated images captured by the camera can be employed to establish the authenticity of digital color images. Digital watermarking (I.J. Cox & M.L. Miller & J.A. Bloom, 2002; H. Liu & J. Rao & X. Yao, 2008) has been proposed as a means to authenticate an image. However, a watermarking must be inserted at the time of recording, which would limit this approach to specially equipped digital cameras having no capabilities to add a watermarking at the time of image capture. Furthermore, the watermarking would be destroyed if the image is compressed and the ruin of watermark would make the method failed. Passive (nonintrusive) image forensics is regarded as the future direction. In contrast to the active methods, blind approaches need no prior information that is used in the absence of any digital watermarking or signature. Blind approaches can be roughly grouped into five categories (Farid, 2009):

(1) pixel-based techniques that analyze pixel-level correlations arising from tampering. Efficient algorithms based on pixels have been proposed to detect cloned (B. Mahdian & S. Saic, 2007; A. Popescu & H. Farid, 2004; J. Fridrich & D. Soukal & J. Lukas, 2003), re-sampled (A. C. Popescu & H. Farid, 2005), spliced (T. T. Ng & S. F. Chang, 2004; T. T. Ng & S. F. Chang & Q. Sun, 2004; W. Chen, & Y. Shi, & W. Su, 2007) images.Statistical properties (H. Farid & S. Lyu, 2003; S. Bayram, & N. Memon, & M. Ramkumar, & B. Sankur, 2004) in natural images are also utilized;

(2) format-based techniques detect tampering in lossy image compression: unique properties of lossy compression such as JPEG can be exploited for forensic analysis (H. Farid, 2008; J. Lukas & J. Fridrich, 2003; T. Pevny & J. Fridrich, 2008).

(3) camera-based techniques exploit artifacts introduced by the camera lens, sensor or on-chip post-processing (J. Lukas, & J. Fridrich & M. Goljan, 2005; A. Swaminathan & M. Wu & K. J. Ray Liu, 2008). Models of color filter array (A. C. Popescu & H. Farid, 2005; S. Bayram & H. T. Sencar & N. Memon, 2005), camera response (Y. F. Hsu & S. F. Chang, 2007; Z. Lin & R. Wang & X. Tang & H.Y. Shum, 2005) and sensor noise (H. Gou & A. Swaminathan & M. Wu, 2007; M. Chen & J. Fridrich &M. Goljan & J. Lukas ,2008; J. Lukas, & J. Fridrich & M. Goljan, 2005) are estimated to infer the source digital cameras and reveal digitally altered images. Other work such as (A. Swaminathan & M. Wu & K. J. Ray Liu, 2008) trace the entire in-camera and post-camera processing operations to identify the source digital cameras and reveal digitally altered images using the intrinsic traces.

(4) physically-based techniques model and detect anomalies using physical rules. For example, three dimensional interaction between physical objects, light, and the camera can be used as evidence of tampering (M.K. Johnson & H. Farid, 2005; M. K. Johnson & H. Farid, 2007).

(5) geometric-based techniques make use of geometric constraints that are preserved or recovered from perspective views (M. K. Johnson & H. Farid, 2006; M. K. Johnson, 2007; W. Wang & H. Farid, 2008; W. Zhang & X. Cao & Z. Feng & J. Zhang & P. Wang, 2009; W. Zhang & X. Cao & J. Zhang & J.Zhu.&P. Wang, 2009).

Several geometric-based techniques (M. K. Johnson & H. Farid, 2007; W. Wang & H. Farid, 2008; W. Zhang & X. Cao & Z. Feng & J. Zhang & P. Wang, 2009; M. K. Johnson & H. Farid, 2006) have been proposed in the field of image forgery detection. The estimation of internal camera parameters including principal point (M. K. Johnson & H. Farid, 2007) and skew (W. Wang & H. Farid, 2008) can be used as evidence of tampering. In (M. K. Johnson & H. Farid, 2007) the authors showed how translation in the image plane is equivalent to a shift of the principal point and differences in which can therefore be used as evidence of forgery. Wang and Farid (W. Wang & H. Farid, 2008) argued that the skew of the re-projected video is inconsistent with the expected parameter of an authentic video. The approach has the advantage that the re-projection can cause a non-zero skew in the camera intrinsic parameters, but there are also some drawbacks that it only applies to frames that contain a planar surface. Zhang et al. (W. Zhang & X. Cao & Z. Feng & J. Zhang & P. Wang, 2009) described a technique for detecting image composites by enforcing two-view geometrical constraints. The approach can detect fake regions efficiently on pictures at the same scene but requires two images correlated with **H** (planar homography) or **F** (fundamental matrix) constraints.

Metric measurements can be made from a planar surface after rectifying the image. In (M. K. Johnson & H. Farid, 2006), the authors reviewed three techniques for the rectification of planar surfaces under perspective projection. They argued that knowledge of polygons of known shape, two or more vanishing points, and two or more coplanar circles can be used to recover the image to world transformation of the planar surface, thereby allowing metric measurements to be achieved on the plane. Each method in (M. K. Johnson & H. Farid, 2006) requires only one single image but fails in measurements for objects out of the reference plane. Wang et al. (G. Wang & Z. Hu & F. Wu & H. T. Tsui, 2005) show how to use the camera matrix and some available scene constraints to retrieve geometrical entities of the scene, such as height of an object on the reference plane, measurements on a vertical or arbitrary plane with respect to the reference plane, etc.

The single view metrology using geometric constraints has been addressed in (A. Criminisi & I. Reid & A. Zisserman, 1999). The authors demonstrated that the affine 3D geometry of a scene may be measured from a single perspective image using the vanishing line of a reference plane and the vertical vanishing point. However, they are only concerned with measurements of the distance between the plane which is parallel to the reference plane and measurements on this plane.

This chapter is organized as follows. After reviewing the background in section 2, the involved methods based on geometric technique are described in sections 3. The future research direction is given in section 4 and the final conclusions are drawn in section 5.

## BACKGROUND

Photographic alterations have existed about as long as photography itself. However, before the digital age, such deceptions required mastery of complex and time-consuming darkroom techniques. Nowadays, anyone who has a little of computer skill can use powerful and inexpensive editing software to create tampered images as he or she likes. Therefore, as sophisticated forgeries appear with fast and alarming frequency, people's belief in what they see has been eroded (H. Farid, 2009).

A more recent example of photo tampering came to light in July 2008. Sepah News, the media arm of Iran's Revolutionary Guard, celebrated the country's military prowess by releasing a photo showing the simultaneous launch of four missiles. But only three of those rockets actually left the ground, a fourth was digitally added. The truth emerged after Sepah circulated the original photo showing three missiles in flight—but not before the faked image appeared on the front pages of the Chicago Tribune, the Financial Times, and the Los Angeles Times.

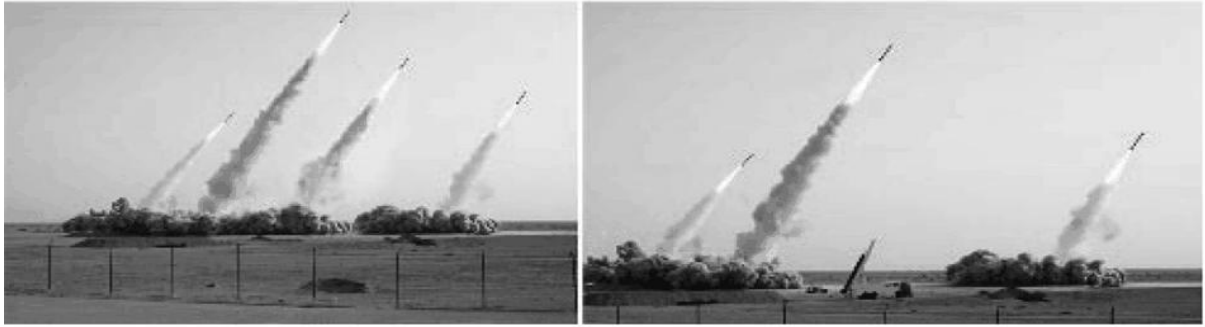

*Figure 1. A July 2008 photo shows four Iranian missiles streaking skyward. The right is the true image Sepah News replaced the faux photo with the original without explanation.*

Over the past few years, the field of digital-image forensics has emerged to challenge this growing problem and return some level of trust in photographs. Nearly every digital forgery starts out as a photo taken by a digital camera. The camera's image sensor acts as the film. By using computer methods to look at the underlying patterns of pixels that make up a digital image, specialists can detect the often-subtle signatures of manipulated images that are invisible to the naked eye.

Traditionally, watermarking is added into the images or video to give the validating information for image authentication. However, the watermarking can be easily destroyed in the process of image compression. Recently, digital blind techniques emerge in the field of image forgery detection. These techniques work on the assumption that although digital forgeries may leave no visual clues that indicate tampering, they may alter the underlying statistics of an image.

## MAIN FOCUS AND CONTRIBUTION OF THE CHAPTER

In this chapter, we focus on the category on the geometric-based techniques in image forgery detection. Geometric techniques, which appear as a new application: the nonintrusive digital image forensic can be further divided into four categories: (1) techniques based on the camera's intrinsic parameters; (2) techniques based on metric measurement; (3) techniques based on multiple view geometry; (4) techniques based on other geometrical constraints.

After reviewing the literature in the field, we propose a potential solution for image integrity's authentication, which is based on the published geometric method on 3D height measurement, measurements on the vertical or an arbitrary plane with respect to a reference plane. Our proposed solution enriches the detecting methods for image forgery, which provides a prospect to build a integrated framework incorporating various methods for fake region detection.

### Solutions and Recommendations

**Preliminary**

**Camera model** The general pinhole camera can also be written as:

$$\mathbf{P} = [\mathbf{r_1} \quad \mathbf{r_2} \quad \mathbf{r_3} \quad \mathbf{t}], \mathbf{K} = \begin{bmatrix} f & \gamma & u_0 \\ 0 & \lambda f & v_0 \\ 0 & 0 & 1 \end{bmatrix} \quad (1)$$

where $\mathbf{r}_i$ is the $i^{th}$ column of the rotation matrix $\mathbf{R}$, $\mathbf{t}$ is the translation vector, and $\mathbf{K}$ is a non-singular 3×3 upper triangular matrix known as the camera calibration matrix including five parameters, i.e. the focal length $f$, the skew $\gamma$, the aspect ratio $\lambda$ and the principal point at $(u_0, v_0)$.

**Planar homography** Suppose there is a plane in the scene, without loss of generality, we define the origin of the coordinate frame lie on this plane (i.e. reference plane), with the X and Y- axes spanning the plane. The Z-axis is the reference direction, which is any direction not parallel to the plane. A 3D point $\mathbf{M} = [x \quad y \quad 0 \quad w]^T$ and its corresponding image projection $\mathbf{m} = [u \quad v \quad 1]^T$ are related via a 3×4 matrix $\mathbf{P}$ by

$$\mathbf{m} \sim \mathbf{PM} = \mathbf{P}[x \quad y \quad 0 \quad w]^T = \underbrace{[\mathbf{p_1} \quad \mathbf{p_2} \quad \mathbf{p_4}]}_{\mathbf{H}} [x \quad y \quad w]^T = \begin{bmatrix} h_{11} & h_{12} & h_{13} \\ h_{21} & h_{22} & h_{23} \\ h_{31} & h_{32} & h_{33} \end{bmatrix} \begin{bmatrix} x \\ y \\ w \end{bmatrix} \quad (2)$$

where ~ indicates equality up to multiplication by a non-zero scale factor, $h_{ij}$ is the component of $\mathbf{H}$, and $\mathbf{p}_i$ is the $i^{th}$ column of $\mathbf{P}$. Hence, the projection from a point on the plane to its image is simplified as

$$\mathbf{m} \sim \mathbf{HM}' \quad (3)$$

where, $\mathbf{H} = [\mathbf{p_1} \quad \mathbf{p_2} \quad \mathbf{p_4}]$ is called plane to plane homography, $\mathbf{M}' = [x \quad y \quad w]^T$ is a homogeneous vector for a point on the reference plane. Usually, $\mathbf{H}$ is a 3×3 homogeneous matrix with 8 degrees of freedom (*dof*).

Here, for generality, we introduce an approach to estimate the transformation from 2D points in image to metric rectification of world coordinates up to similarity's ambiguity. This transformation, $\mathbf{H}$, can be decomposed into the multiplication of two matrices:

$$\mathbf{H} = \mathbf{H_A}\mathbf{H_P} \quad (4)$$

where $\mathbf{H_A}$ and $\mathbf{H_P}$ represent affine and pure projective transformations respectively:

$$\mathbf{H_P} = \begin{pmatrix} 1 & 0 & 0 \\ 0 & 1 & 0 \\ l_1 & l_2 & l_3 \end{pmatrix} \quad (5)$$

$$\mathbf{H_A} = \begin{pmatrix} \frac{1}{\beta} & \frac{\alpha}{\beta} & 0 \\ 0 & 1 & 0 \\ 0 & 0 & 1 \end{pmatrix} \quad (6)$$

where $\mathbf{l}_\infty = (l_1, l_2, l_3)^T$ is the vanishing line of the reference plane that is determined by two vanishing points, the coefficients $\alpha$ and $\beta$ are estimated as follows. Given a known angle $\theta$ on the world plane between two lines $\mathbf{m} = (m_1, m_2, m_3)^T$ and $\mathbf{n} = (n_1, n_2, n_3)^T$ (parameterized as homogeneous vectors), it can be shown in (D. Liebowitz & A. Zisserman, 1998) that $\alpha$ and $\beta$ lie on a circle with center:

$$(c_\alpha, c_\beta) = (\frac{a+b}{2}, \frac{a-b}{2}\cot(\theta)) \quad (7)$$

with radius:

$$r = \left| \frac{(a-b)}{2\sin(\theta)} \right| \quad (8)$$

where $a = -m_2/m_1, b = -n_2/n_1$.

Alternatively, two equal but unknown angles on the world plane between two lines imaged with directions $a_1, b_1$ and $a_2, b_2$ also provide a constraint circle with center:

$$(c_\alpha, c_\beta) = (\frac{a_1 b_2 - b_1 a_2}{a_1 - b_1 - a_2 + b_2}, 0) \tag{9}$$

and radius:

$$r^2 = (\frac{a_1 b_2 - b_1 a_2}{a_1 - b_1 - a_2 + b_2})^2 + \frac{(a_1 - b_1)(a_1 b_1 - a_2 b_2)}{a_1 - b_1 - a_2 + b_2} \tag{10}$$

Thereby, two pairs of equal but unknown angles can provide two circles, and $\alpha$, $\beta$ will be obtained from the intersection of such two circles.

Besides, a known length ratio between two non-parallel line segments **i** and **j** on the world plane can provide another constraint circle with center:

$$(c_\alpha, c_\beta) = (\frac{\Delta i_1 \Delta i_2 - \rho^2 \Delta j_1 \Delta j_2}{\Delta i_2^2 - \rho \Delta j_2^2}, 0) \tag{11}$$

and with radius:

$$r = \left| \frac{\rho(\Delta j_1 \Delta i_2 - \Delta i_1 \Delta j_2)}{\Delta i_2^2 - \rho \Delta j_2^2} \right| \tag{12}$$

where $\rho$ is the known length ratio between **i** and **j**, and $\Delta i_1 \Delta i_2$ are the differences between the first and second coordinates of the line segment **i**. We can also use another combination: length ratio and a known angle to obtain the affine transformation $\mathbf{H_A}$.

**Fundamental matrix** The fundamental matrix **F** encapsulates the intrinsic projective geometry between two views which only depends on the camera's internal parameters and relative pose. F is a 3×3 matrix of rank 2. If a point in 3-space is imaged as x in the first view, and $\mathbf{x}'$ in the second, then the image points satisfy the relation $\mathbf{x}'\mathbf{F}\mathbf{x} = 0$.

If points x and $\mathbf{x}'$ correspond, then $\mathbf{x}'$ lies on the epipolar line $\mathbf{l}' = \mathbf{F}\mathbf{x}$ corresponding to the point x. In other words $\mathbf{x}'\mathbf{F}\mathbf{x} = \mathbf{x}'^T\mathbf{l} = 0$.

## Solutions
### Image forgery detection based on camera internal parameters

The internal parameters in the camera matrix (Eq. (1)) are focal length, skew, aspect ratio and principal point. The estimated internal parameters recovered from a non-tampered image should be inconsistent across the image. Therefore, Differences in these parameters across the image are used as evidence of tampering

**Using principal point** It is a common thing that a single image is composited of two or more people, for example, figure 2, is a composite of actress Marilyn Monroe (1926-1962) and President Abraham Lincoln (1809-1865).

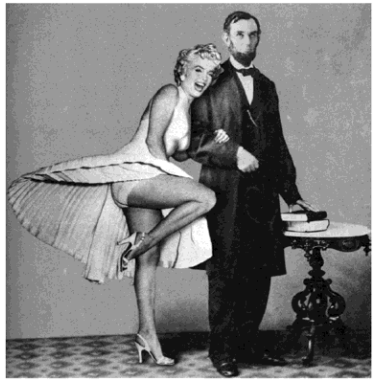

*Figure 2. Composite of Marilyn Monroe and Abraham Lincoln.*

Estimating a camera's principal point from the image of a person's eyes is a feasible approach to authenticate a image's integrity. Inconsistencies in the principal point can be used as evidence of tampering. In authentic images, the principal point is near the center of the image. In (M. K. Johnson & H.

Farid, 2007), the authors stated that the principal point is moved proportionally when a person is translated in the image as part of creating a composite.

This section first describes how the planar homography $\mathbf{H}$ can be estimated from an image of a person's eyes and show how this transform can be factored into a product of matrices that embody the camera's intrinsic and extrinsic parameters. Then it will be shown how translation in the image plane can be detected from inconsistencies in the estimated camera's intrinsic parameters.

Since the world points lie on a single plane, $\mathbf{H}$ can be decomposed in terms of the intrinsic and extrinsic parameters:

$$\mathbf{H} = [\mathbf{h}_1 \quad \mathbf{h}_2 \quad \mathbf{h}_3] = s\mathbf{K}(\mathbf{r}_1 \quad \mathbf{r}_2 \quad \mathbf{t}) \tag{13}$$

where s is a scale factor and $\mathbf{K}$ is the camera calibration matrix. For simplicity, it is assumed that the skew is zero and the aspect ratio is 1. Under these assumptions, the matrix $\mathbf{K}$ is:

$$\mathbf{K} = \begin{bmatrix} f & 0 & u_0 \\ 0 & f & v_0 \\ 0 & 0 & 1 \end{bmatrix} \tag{14}$$

The camera's intrinsic components can be estimated by decomposing $\mathbf{H}$ according to Equation (13). It is evident to show that $\mathbf{r}_1 = \frac{1}{s}\mathbf{K}^{-1}\mathbf{h}_1$ and $\mathbf{r}_2 = \frac{1}{s}\mathbf{K}^{-1}\mathbf{h}_2$. The constraints that are orthogonal and have the same norm yield two constraints on the matrix K:

$$\mathbf{r}_1^T \mathbf{r}_2 = \mathbf{h}_1^T (\mathbf{K}^{-T}\mathbf{K}^{-1})\mathbf{h}_2 = 0 \tag{15}$$

$$\mathbf{r}_1^T \mathbf{r}_1 - \mathbf{r}_2^T \mathbf{r}_2 = \mathbf{h}_1^T (\mathbf{K}^{-T}\mathbf{K}^{-1})\mathbf{h}_1 - \mathbf{h}_2^T (\mathbf{K}^{-T}\mathbf{K}^{-1})\mathbf{h}_2 = 0 \tag{16}$$

It is possible to estimate the principal point $(u_0, v_0)$ or the focal length $f$, but not the both with only two constraints. As such, the authors assumed a known focal length. Therefore, the homography H can be achieved from image of people's eyes.

The translation of two circles (eyes) in the image is equivalent to translating the camera's principal point. In homogeneous coordinates, translations are represented by multiplication with a translation matrix T:

$$\mathbf{y} = \mathbf{T}\mathbf{x} \tag{17}$$

where:

$$\mathbf{T} = \begin{pmatrix} 1 & 0 & d_1 \\ 0 & 1 & d_2 \\ 0 & 0 & 1 \end{pmatrix} \tag{18}$$

and the amount of translation is $(d_1, d_2)$. The mapping from world X to image coordinates y is:

$$\mathbf{y} = \mathbf{T}\mathbf{H}\mathbf{x} = s\mathbf{T}\mathbf{K}(\mathbf{r}_1 \quad \mathbf{r}_2 \quad \mathbf{t})\mathbf{X} = s\hat{\mathbf{K}}(\mathbf{r}_1 \quad \mathbf{r}_2 \quad \mathbf{t})\mathbf{X} \tag{19}$$

where

$$\hat{\mathbf{K}} = \begin{pmatrix} f & 0 & u_0 + d_1 \\ 0 & f & v_0 + d_2 \\ 0 & 0 & 1 \end{pmatrix} \tag{20}$$

Therefore, translation in image coordinates is equivalent to translating the principal point. If the principal point in an authentic image is near the origin which has large deviations from the image center, or inconsistencies in the estimated principal point across the image, can be used as the evidence of tampering.

**Using skew** Another camera internal parameter, the skew, can also be used as the evidence of tampering. In figure 3, the right is a re-projected frame of a movie, the left is the same scene as viewed on a movie screen.

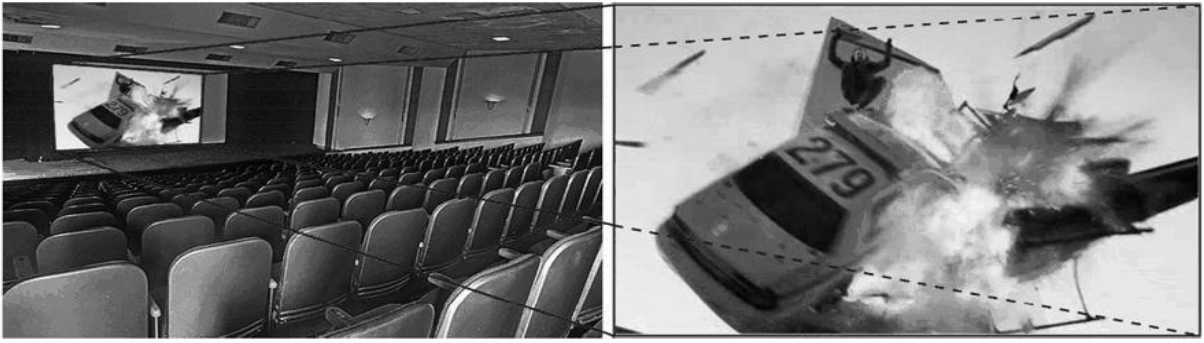

*Figure 3. Projected movie introduces distortions that can be used to detect re-projected video.*

The author in (W. Wang & H. Farid, 2008) described an automatic technique for detecting a video that was recorded from a screen. It is ready to show that the internal camera parameters of such video are inconsistent with the expected parameters of an authentic video. Due to the angle of the video camera relative to the screen in the re-projected process, a perspective distortion has been introduced into this second recording. It can introduce a distortion into the intrinsic camera parameters the camera skew which depends on the angle between the horizontal and vertical pixel axes. In (W. Wang & H. Farid, 2008), the authors demonstrated that re-projection can cause a non-zero skew in the camera's intrinsic parameters. Now we review two approaches for estimating camera skew from a video sequence.

**Skew estimation I** The projection of a planar surface is given by:
$$\mathbf{y} = s\mathbf{KPY} = s\mathbf{HY} \tag{21}$$

where $\mathbf{K}$ and $\mathbf{P}$ are the intrinsic and extrinsic matrices, $\mathbf{y}$ is the 2-D projected point in homogeneous coordinates, and $\mathbf{Y}$, in the appropriate coordinate system, is specified by 2-D coordinates in homogeneous coordinates. The $3 \times 3$ matrix $\mathbf{H}$ is a non-singular matrix referred to as a homography. Given the above equality, the left- and right-hand sides of this homography satisfy the following:

$$\mathbf{y} \times (\mathbf{HY}) = 0$$
$$\begin{pmatrix} y_1 \\ y_2 \\ y_3 \end{pmatrix} \left( \begin{pmatrix} h_{11} & h_{21} & h_{31} \\ h_{12} & h_{22} & h_{32} \\ h_{13} & h_{23} & h_{33} \end{pmatrix} \begin{pmatrix} Y_1 \\ Y_2 \\ Y_3 \end{pmatrix} \right) = 0 \tag{22}$$

A matched set of points y and Y appear to provide three constraints on the eight unknown elements of H which is defined only up to an unknown scale factor.

Next we describe how to estimate the camera skew from the estimated homography $\mathbf{H}$. Since $\mathbf{H}$ can be expressed as:
$$\mathbf{H} = \mathbf{KP} = \mathbf{K}(\mathbf{p}_1 \quad \mathbf{p}_2 \mid \mathbf{t}) \tag{23}$$

The orthonormality of $\mathbf{p}_1$ and $\mathbf{p}_2$, yields the following two constraints:
$$\mathbf{p}_1^T \mathbf{p}_2 = 0 \text{ and } \mathbf{p}_1^T \mathbf{p}_1 = \mathbf{p}_2^T \mathbf{p}_2 \tag{24}$$

which in turn imposes the following constraints on H and K:

$$\begin{pmatrix} h_{11} \\ h_{12} \\ h_{13} \end{pmatrix}^T K^{-T} K^{-1} \begin{pmatrix} h_{21} \\ h_{22} \\ h_{23} \end{pmatrix} = 0 \tag{25}$$

$$\begin{pmatrix} h_{11} \\ h_{12} \\ h_{13} \end{pmatrix}^T K^{-T} K^{-1} \begin{pmatrix} h_{11} \\ h_{12} \\ h_{13} \end{pmatrix} = \begin{pmatrix} h_{21} \\ h_{22} \\ h_{23} \end{pmatrix}^T K^{-T} K^{-1} \begin{pmatrix} h_{21} \\ h_{22} \\ h_{23} \end{pmatrix} \tag{26}$$

Denote $\mathbf{B} = \mathbf{K}^{-T}\mathbf{K}^{-1}$, where B is a symmetric matrix parameterized with three degrees of freedom:

$$\mathbf{B} = \begin{pmatrix} b_{11} & b_{12} & 0 \\ b_{12} & b_{22} & 0 \\ 0 & 0 & 1 \end{pmatrix} \quad (27)$$

Each image of a planar surface enforces two constraints on the three unknowns $b_{ij}$. The matrix $\mathbf{B} = \mathbf{K}^{-T}\mathbf{K}^{-1}$ can, therefore, be estimated from two or more views of the same planar surface using standard least-square estimation. The desired skew can be then determined from the estimated matrix B as:

$$s = -f \frac{b_{12}}{b_{11}} \quad (28)$$

**Skew estimation II** Here we review an approach that does not require any known world geometry, but requires a non-linear minimization. Suppose two frames of a video sequence with corresponding image points given by u and v. The corresponding points satisfy the following relationship:

$$\mathbf{v}^T \mathbf{F} \mathbf{u} = 0 \quad (29)$$

The above relationship can yield:

$$(v_1 \quad v_2 \quad 1) \begin{pmatrix} f_{11} & f_{21} & f_{31} \\ f_{12} & f_{22} & f_{32} \\ f_{13} & f_{23} & f_{33} \end{pmatrix} \begin{pmatrix} u_1 \\ u_2 \\ 1 \end{pmatrix} = 0 \quad (30)$$

Each pair of matched points u and v provides one constraint for the eight unknown elements. Therefore, eight or more matched pairs of points are required to solve for the components of the fundamental matrix. It then will be described how to estimate the camera skew from the estimated fundamental matrix $\mathbf{F}$. Assume the intrinsic camera matrix $\mathbf{K}$, is the same across the views containing the matched image points. The essential matrix E is defined as:

$$\mathbf{E} = \mathbf{K}^T \mathbf{F} \mathbf{K} \quad (31)$$

The essential matrix E has rank 2 and the two non-zero singular values of E are equal. This property will be exploited to estimate the camera skew. The cost function that is minimized in terms of the camera focal length $f$ and skew $s$:

$$C(f,s) = \sum_{i=1}^{n} \frac{\sigma_{i1} - \sigma_{i2}}{\sigma_{i2}} \quad (32)$$

Where $\sigma_{i1}$ and $\sigma_{i2}$ are the non-zero singular values of E from n estimated fundamental matrices, and K is parameterized as:

$$\mathbf{K} = \begin{pmatrix} f & s & 0 \\ 0 & f & 0 \\ 0 & 0 & 1 \end{pmatrix} \quad (33)$$

**Image forgery detection based on metric measurement** Obtaining metric measurement from a single image is proved useful in forensic settings where real-world measurements are required. Three techniques for making metric measurements on planar surfaces from a single image are reviewed here. In (M. K. Johnson & H. Farid, 2006), the authors surveyed three techniques for the rectification of planar surfaces imaged under perspective projection. Each method requires only a single image. Three methods exploit knowledge of polygons of known shape, two or more vanishing points on a plane, two or more coplanar circles, respectively. In each case, the world to image transformation of the planar surface can be recovered, thereby allowing metric measurements to be made on the plane.

**Polygons** Under an ideal pinhole camera model, points on a plane, $\mathbf{X}$, in the world coordinate system are imaged to the image plane with coordinates $\mathbf{x}$, given by:

$$\mathbf{x} = \mathbf{H}\mathbf{X} \quad (34)$$

where both points are homogeneous 3-vectors in their respective coordinate systems. In order to solve for the projective transformation matrix $\mathbf{H}$, four or more points with known coordinates $\mathbf{X}$ and $\mathbf{x}$ are required. The estimation of $\mathbf{H}$ is determined up to an unknown scale factor. From a single image, a known length on the world plane is required to determine this scale factor. With a known **H**, the image is warped according to $\mathbf{H}^{-1}$ to yield a rectified image, from which measurements can be made.

Shown in Figure 4 is a tampered image – two boxes of Marlboro cigarettes were doctored to read "Marlboro kids" with an image of the cartoon character Tweety Bird. The center and right of Figure 4 are the results of planar rectification based on the known shape of the rectangle on the front of the box. It is obvious that after rectification the text and character on the boxes are inconsistent with another, clearly revealing them to be fakes.

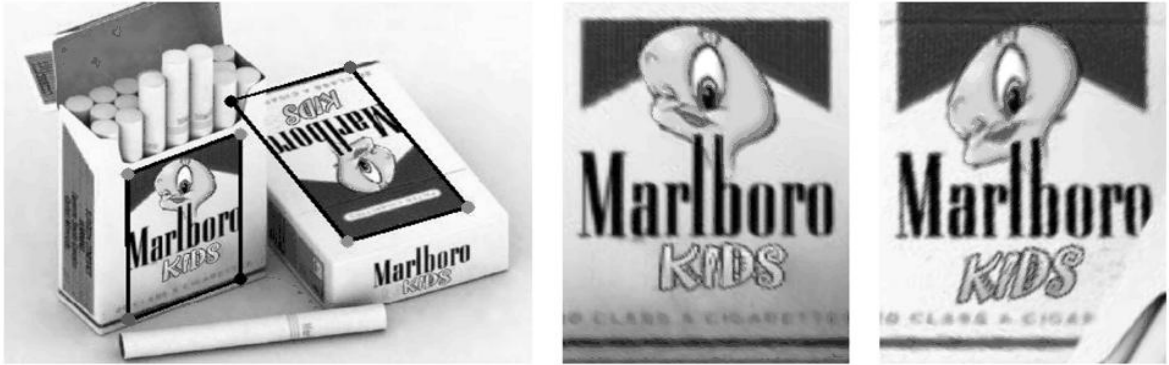

*Figure 4. The face of each cigarette box is rectified using the known shape of the box face. The center and right panels are the rectified images, clearly showing an inconsistency in the cartoon character and text.*

**Vanishing points**   Here, we review how two or more vanishing points can be used to make metric measurements on a planar surface in the world. Consider the inverse mapping:

$$\mathbf{X} = \mathbf{H}^{-1}\mathbf{x} = (\mathbf{H}_s \mathbf{H}_A \mathbf{H}_P)\mathbf{x} \tag{35}$$

The projective transformation matrix $\mathbf{H}^{-1}$ is uniquely decomposed into a product of three matrices: a similarity matrix $\mathbf{H}_S$, an affine matrix $\mathbf{H}_A$ and a pure projective matrix $\mathbf{H}_P$. The final similarity matrix $\mathbf{H}_S$ is given by:

$$\mathbf{H}_s = \begin{pmatrix} sr_1 & sr_2 & t_x \\ sr_3 & sr_4 & t_y \\ 0 & 0 & 1 \end{pmatrix} = \begin{pmatrix} s\mathbf{R} & \mathbf{t} \\ \mathbf{0}^T & 1 \end{pmatrix} \tag{36}$$

where *s* is an isotropic scaling, R is a rotation matrix, and t is a translation vector. Only the scale factor *s* is required in order to make absolute Euclidean measurements on a world plane. From a single image, a known length on the world plane is required in order to determine this scale factor. The image is then warped according to $\mathbf{H}^{-1}$ to yield a rectified image, from which measurements can be made.

Figure 5 is an image of two people standing outside of a store. Also shown in this figure is a rectified version of this image. The lines along the building face were used to find vanishing points. Since the two people are standing in a plane that is approximately parallel to the plane of the store front, their relative heights can be measured after rectification. Using the height of the person on the left as a reference (64.75 inches), the height of the person on the right was estimated to be 69.3 inches. This person's actual height is 68.75 inches, yielding an error of 0.55 inches or 0.8%.

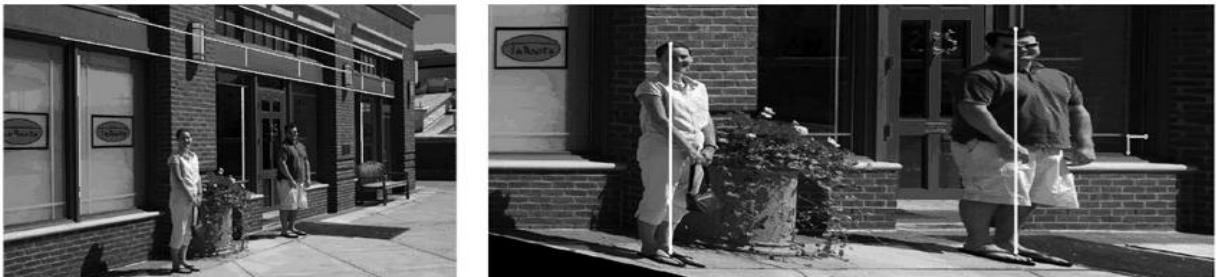

*Figure 5. The wall was rectified using vanishing lines, a known angle and length ratio. The right is the*

*rectified image, from which measurements of the two people can be made.*

**Circles** The circular points $\mathbf{I} = (1 \quad i \quad 0)^T$ and $\mathbf{J} = (1 \quad -i \quad 0)^T$ lie on every circle in a plane. Consider the effect of the projective transform $\mathbf{H}$ on the circular points $\mathbf{I}$ and $\mathbf{J}$. The mapping from world to image coordinates is:

$$\mathbf{H} = (\mathbf{H_S H_A H_P})^{-1} = \mathbf{H_P^{-1} H_A^{-1} H_S^{-1}} \tag{37}$$

The circular points are mapped under this projective transform $\mathbf{H}$: $\mathbf{HI}$ and $\mathbf{HJ}$. The similarity matrix $\mathbf{H_S^{-1}}$ can be ignored because the circular points are either invariant or are swapped under this transformation. Therefore, without loss of generality, it can be assumed that the circular points are invariant to $\mathbf{H_S^{-1}}$, in which:

$$\mathbf{HI} = \mathbf{H_P^{-1} H_A^{-1} I} = \mathbf{H_P^{-1}} \begin{pmatrix} \beta + i\alpha \\ i \\ 0 \end{pmatrix} = \begin{pmatrix} \beta + i\alpha \\ i \\ -l_1/l_3(\beta + i\alpha) - il_2/l_3 \end{pmatrix} \tag{38}$$

The projection of $\mathbf{J}$ is the complex conjugate of $\mathbf{HI}$. $\mathbf{HI}$ and $\mathbf{HJ}$ contain the required coefficients of the desired matrices $\mathbf{H}_P$ and $\mathbf{H}_A$. Under the projective transform, the intersections of any two coplanar circles are preserved, while these circles are projected to ellipse in the image. As a result, it is important to compute the intersection of these ellipses, two of which correspond to $\mathbf{HI}$ and $\mathbf{HJ}$.

**Image forgery detection based on multiple view geometry** W. Zhang & X. Cao & Z. Feng & J. Zhang & P. Wang(2009) stated a technique for detecting image composites by enforcing two-view geometrical constrains: $\mathbf{H}$ and $\mathbf{F}$ constraints on image pairs, where $\mathbf{H}$ denotes the planar homography matrix and $\mathbf{F}$ the fundamental matrix.

**H constraint** When a camera rotates for an angle, corresponding points $\mathbf{x}_1$ and $\mathbf{x}_2$ on two image planes are related by:

$$\mathbf{x}_2 = \mathbf{K}[\mathbf{R}|0]\mathbf{X} = \mathbf{KRK}^{-1}\mathbf{x}_1 \tag{37}$$

While in case of pure zooming, corresponding points hold:

$$\mathbf{x}_2 = \mathbf{K}'[\mathbf{R}|0]\mathbf{X} = \mathbf{K}'\mathbf{K}^{-1}\mathbf{x}_1 \tag{38}$$

Where $\mathbf{K}$ and $\mathbf{K}'$ are two internal parameter matrices.

Pictures taken before and after camera motion are constrained by a planar homography, $\mathbf{H}$, if any of the assumptions holds: (1) camera does not change its position; (2) scenes viewed are locally coplanar.

**F constraint** When a camera moves generally and points are not coplanar, they still can be related with a Fundamental Matrix, $\mathbf{F}$, which maps a point $\mathbf{x}_1$ on one image to a line $\mathbf{l}_1$ on the other image:

$$\mathbf{x}_2^T \mathbf{l}_1 = \mathbf{x}_2^T \mathbf{F} \mathbf{x}_1 = 0 \tag{39}$$

Where $\mathbf{x}_2$ is the corresponding point of $\mathbf{x}_1$.

For $\mathbf{H}$ constraint, the authors combine bucketing technique and RANSAC for the estimation of $\mathbf{H}$, since SIFT fails to find initial matches for the estimation of $\mathbf{H}$ and $\mathbf{F}$. After the estimation of H, the rectified image $\mathbf{I}_2$ can be recovered from the original image $\mathbf{I}_1$. The difference matrix $\mathbf{D}$ is introduced to evaluate the similarity of the image $\mathbf{I}_1$ and $\mathbf{I}_2$. Regions with high difference $\mathbf{D}$ are selected to produce the binary map, which highlights fake regions. The threshold of cutting the difference is given by:

$$t = \max(\mathbf{D}) - c \tag{40}$$

where $\mathbf{D}$ denotes the difference of image $\mathbf{I}_1$ and $\mathbf{I}_2$, at every pixel, and the constant value c locates in [0.3, 0.6].

For detecting composites using $\mathbf{F}$ constraint, "Gold Standard" algorithm (R. Hartley & A. Zisserman, 2004) is applied to estimate $\mathbf{F}$. The point $\mathbf{x}_1$ from frame $\mathbf{I}_1$ maps to an epipolar line on image $\mathbf{I}_2$. Distance between $\mathbf{x}_2$ and epipolar line $\mathbf{l} = \mathbf{F}\mathbf{x}_1$ is used as the metric,

$$d(\mathbf{x}_2, \mathbf{F}\mathbf{x}_1) = \sqrt{(\mathbf{x}_2^T \mathbf{F}\mathbf{x}_1)^2 / ((\mathbf{F}\mathbf{x}_1)_x^2 + (\mathbf{F}\mathbf{x}_1)_y^2)} \quad (41)$$

The distance measurement provides a candidate set, $\psi = \{(\mathbf{x}_{1i}, \mathbf{x}_{2i}) \mid d(\mathbf{x}_{2i}, \mathbf{F}\mathbf{x}_{1i}) > t\}$, of features inside the potential fake regions. Morphological operation is used to dilate the points in $\psi$ to highlight a region including dense fake points.

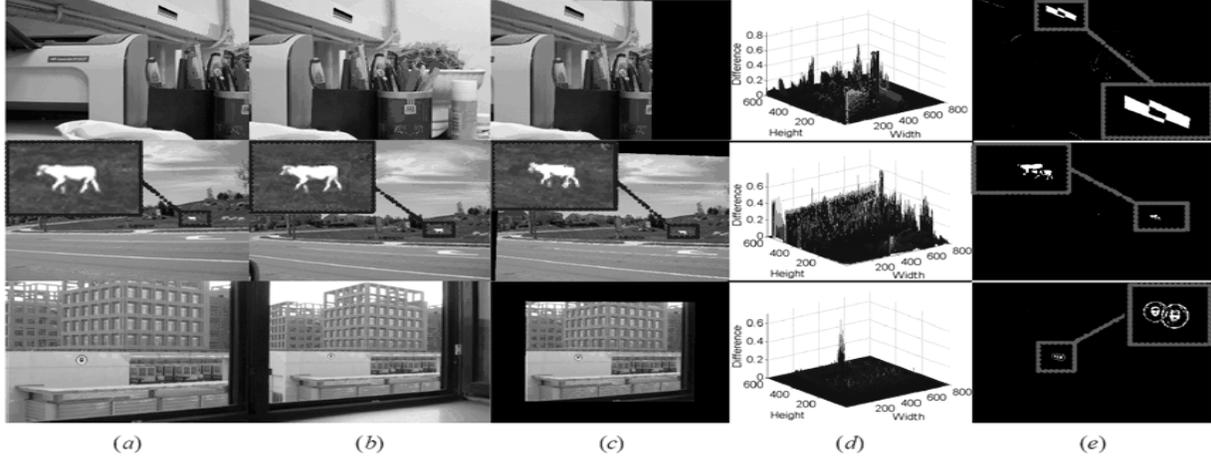

Figure 6. Detecting composites by enforing the **H** constraint. (a, b) Original image pairs. (c) Images rectified from (a) using the estimated **H**. (d) Difference maps between (b) and (c) based on correlation. (e) Binary masks with fake regions. Rows 1 and 2 are images taken with rotation, and row 3 is a zooming case.

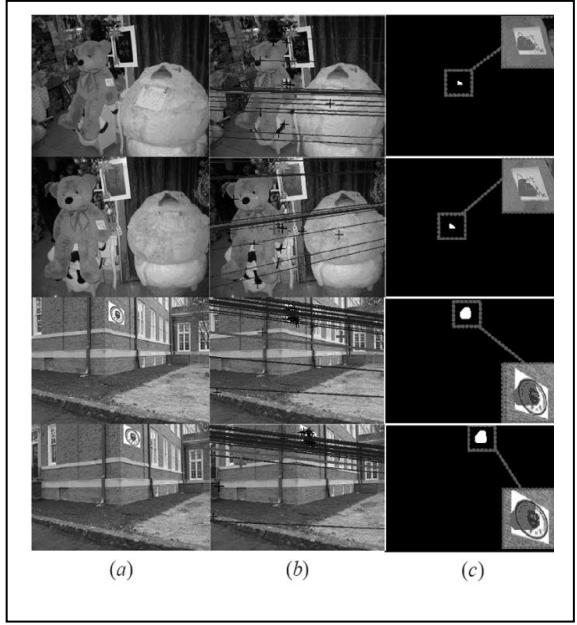

Figure 7. Detecting composites by enforcing the F constraint. (a) Two natural image pairs with visually plausible fake regions. (b) Points with large distance to their corresponding epipolar lines. Some examples are highlighted with larger size. (c) Binary map highlighting the fake regions.

### Image forgery detection using other geometrical constraints

**Shadow geometry** Shadow compositing must be taken into account in some target scenes when there presents shadow projection from the sun. In (W. Zhang & X. Cao & J. Zhang & J.Zhu.&P. Wang,2009) the authors utilize the planar homology (C. E. Springer, 1964) that encompasses the imaged shadow relationship as shown in Figure 7 to detect photo composites. Note that the light source is not necessarily to be at infinity to keep the model by a planar homology, provided that the light source is a point light source, i.e. all light rays are concurrent.

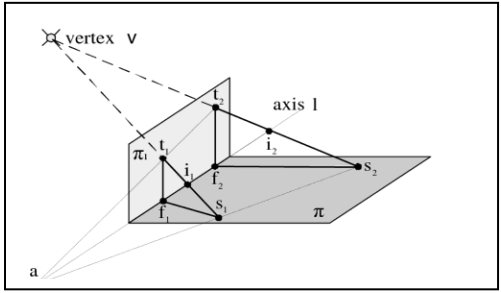

Figure 8. Geometry of a planar homology. A plane $\pi_1$, and its shadow, illuminated by a point light source **v** and cast on a ground plane $\pi$, are related by a planar homology

As shown in Figure 8, a planar homology is a planar projective transformation **H** which has a line **l** of fixed points, called the *axis*, and a distinct fixed point *v*, not on the axis **l**, called the *vertex* of the homology, **H**,

$$\mathbf{H} = \mathbf{I} + (\mu - 1)\frac{\mathbf{v}\mathbf{l}^T}{\mathbf{v}^T\mathbf{l}} \qquad (42)$$

where $\alpha$ is the cross ratio that will be discussed later. In our case, the vertex **v** is the image of the light source, and the axis, **l**, is the image of the intersection between planes $\pi_1$ and $\pi$. Each point off the axis, e.g. $\mathbf{t}_2$, lies on a fixed line $\mathbf{t}_2\mathbf{s}_2$ through **v** intersecting the axis at $\mathbf{i}_2$ and is mapped to another point $\mathbf{s}_2$ on the line. Note that $\mathbf{i}_2$ is the intersection in the image plane, although the light ray $\mathbf{t}_2\mathbf{s}_2$ and the axis, **l**, are unlikely to intersect in 3D real world.

One important property of a planar homology is that the corresponding lines intersect with the axis, e.g. the lines $\mathbf{t}_1\mathbf{t}_2$ and $\mathbf{s}_1\mathbf{s}_2$ intersect at **a** on **l**. Another important property of a planar homology is that the cross ratio, $\mu$, defined by the vertex, **v**, the corresponding points, $\mathbf{t}_i$ and $\mathbf{s}_i$, and the intersection point, $\mathbf{i}_i$, is the characteristic invariance of the homology, and thus is the same for all corresponding points. For example, the cross ratios

$\{\mathbf{v},\mathbf{t}_1;\mathbf{s}_1,\mathbf{i}_1\}$ and $\{\mathbf{v},\mathbf{t}_2;\mathbf{s}_2,\mathbf{i}_2\}$ are equal. The two constraints can be expressed as

$$((\mathbf{t}_2 \times \mathbf{t}_1) \times (\mathbf{s}_2 \times \mathbf{s}_1)) \cdot (\mathbf{f}_2 \times \mathbf{f}_1) = 0 \qquad (43)$$
$$\{\mathbf{v},\mathbf{t}_1;\mathbf{s}_1,\mathbf{i}_1\} = \{\mathbf{v},\mathbf{t}_2;\mathbf{s}_2,\mathbf{i}_2\} \qquad (44)$$

where

$$\mathbf{v} = (\mathbf{t}_2 \times \mathbf{s}_2) \times (\mathbf{t}_1 \times \mathbf{s}_1) \qquad (45)$$

Therefore, these two constraints can be used to detect composites in a nature image. Notice that $\mathbf{t}_1, \mathbf{f}_1, \mathbf{t}_2, \mathbf{f}_2$ have to be coplanar and $\mathbf{f}_1, \mathbf{f}_2$ on the intersection of plane $\pi_1$ and $\pi$. In real world, vertical objects standing on the ground satisfy this assumption, such as standing people, street lamps, trees and buildings. In addition, people usually are interested in inserting a new actor, which is mostly standing and vertical, into some target scene.

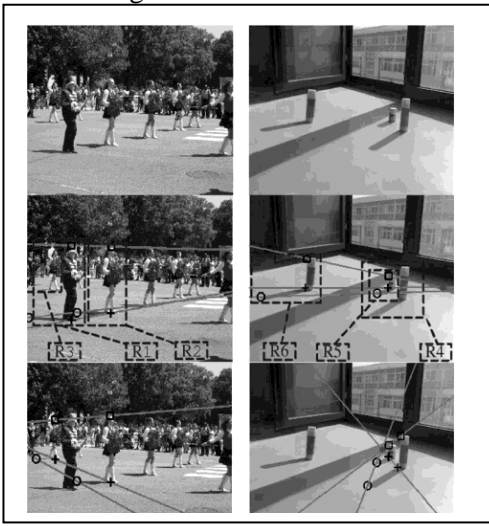

Figure 9. Composite detection based on shadow geometry. **Top row**: Two nature images with composited regions. **Middle row**: The corresponding lines that involve composited regions (R1 and R6) don't intersect on the axis. In addition, they dissatisfy the characteristic invariance constraint in Eq. (44) (see Table 1). **Bottom row**: The imaged shadow relationship of authentic objects can be modeled by a planar homology. Black squares, crosses and circles denote the locations of *t*, *f* and *s* in Figure 8 respectively.

*Table 1. Cross ratios of planar homologies in Figure 9.*

| Region A | Region B | $\mu_A$ | $\mu_B$ | Diff Ratio |
|----------|----------|---------|---------|------------|
| R1 | R2 | 0.1741 | 0.1231 | 29.2589% |
| R2 | R3 | 0.1587 | 0.1573 | 0.8794% |
| R1 | R3 | 0.4454 | 0.4966 | 11.5145% |
| R4 | R5 | 0.6298 | 0.6352 | 0.8647% |
| R4 | R6 | 0.4473 | 0.3384 | 24.3526% |
| R5 | R6 | 0.3237 | 0.2625 | 18.9191% |

As another potential solution for tampered region authentication, we can leverage single view geometric constraints (A. Criminisi & I. Reid & A. Zisserman, 1999) to detect composited regions. General methods (A. Criminisi & I. Reid & A. Zisserman, 1999; G. Wang & Z. Hu & F. Wu & H. T. Tsui, 2005) try to compute the camera matrix **P** induced by rigid constraints to achieve metric rectification. However, we can use minimal geometric constraints including known angles, equal but known angles or length ratios to obtain the planar homography **H**. In the following, we will first show that **P** may be derived from **H** and then, we demonstrate that various geometric combinations can derive the planar homography. After metric rectification is implemented, we present the 3D metric measurement and measurements on the vertical plane or arbitrary plane.

**Metric rectification** The camera matrix **P** can be retrieved from **H** up to 3 degrees of freedom's ambiguities since **P** is 11 *dof* while **H** is 8. Typically, the skew $\gamma$ of the COTS camera is zero, which provides one constraint. The remaining two ambiguities can be relieved by the availability of vertical vanishing point (G. Wang & Z. Hu & F. Wu & H. T. Tsui, 2005), which needs restricted scene to provide the vertical direction. Here, the principal point is instead used since it can provide two independent constraints on **P**, and it is known to be approximately at the center of a natural image (M. K. Johnson & H. Farid, 2007; X. Cao & H. Foroosh, 2006). Therefore, in our implementation, the principal point is assumed as the center of the image and **P** can be determined from **H**. When **H** is determined by using the geometric constraints, then the Image of Absolute Conic (IAC) i.e. $\omega$ can be determined by the orthogonal vanishing points satisfying the following relationship:

$$\mathbf{h}_1^T \omega \mathbf{h}_2 = 0 \tag{46}$$

where $\mathbf{h}_i$ is the $i^{th}$ column of **H**, $\omega = (\mathbf{KK}^T)^{-1}$. After $\omega$ is determined, **K** may be computed by decomposing the IAC. Since $\mathbf{h}_1^T = \mathbf{K}\mathbf{r}_1$, $\mathbf{h}_2^T = \mathbf{K}\mathbf{r}_2$ and $\mathbf{p}_3 = \mathbf{K}\mathbf{r}_3 = \mathbf{K}(\mathbf{r}_1 \times \mathbf{r}_2)$. Thereby, the camera matrix **P** is achieved from **H**.

Simply and particularly, $\omega$ may be directly retrieved from the orthogonal relationship provided by two orthogonal vanishing points which may be available from some scene.

$$\mathbf{v}_x^T \omega \mathbf{v}_y = 0 \tag{47}$$

where $\mathbf{v}_x$ and $\mathbf{v}_y$ are two vanishing points which are perpendicular to each other. And then, **H** may be computed from such geometric constraints.

**Height measurement** We wish to measure the object's 3D height which can be treated as the distance between two parallel planes. The distance between scene planes is specified by a base point on the reference plane and top point in the scene (A. Criminisi & I. Reid & A. Zisserman, 1999). The image containing such planes is illustrated in Figure 10.

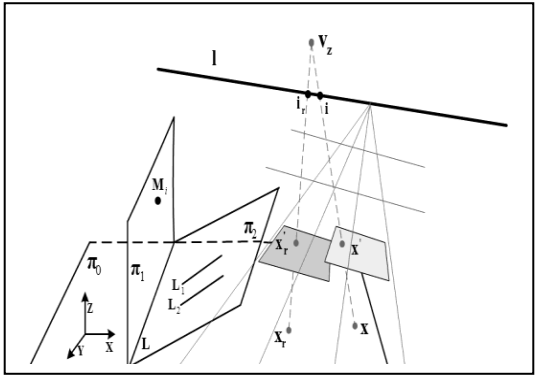

*Figure 10. Height measurement, measurement on vertical plane $\pi_1$ and arbitrary plane $\pi_2$, $\pi_0$ is the reference plane. The point $\mathbf{x_r}$ and $\mathbf{x}$ on the reference plane $\pi_0$ correspond to the point $\mathbf{x_r'}$ and $\mathbf{x'}$. $\mathbf{l}$ is the horizon line, the vertical vanishing point $\mathbf{v_z}$, $\mathbf{i_r}$, $\mathbf{x_r'}$ and $\mathbf{x'}$ can define a cross ratio: $\{\mathbf{v_z}, \mathbf{i_r}; \mathbf{x_r'}, \mathbf{x'}\}$. The same situation holds true for $\{\mathbf{v_z}, \mathbf{i}; \mathbf{x'}, \mathbf{x}\}$. The three planes intersect on line $\mathbf{L}$, and $\mathbf{L_1}$, $\mathbf{L_2}$ are parallel lines on $\pi_2$.*

Suppose the base and the top points can be specified as $\mathbf{X} = [X, Y, 0]^T$ and $\mathbf{X'} = [X, Y, Z]^T$ respectively, and their images are $\mathbf{x}$ and $\mathbf{x'}$. Then the image coordinates are
$\mathbf{x} = \mathbf{PX} = \mathbf{P}[X,Y,0]^T$, $\mathbf{x'} = \mathbf{PX'} = \mathbf{P}[X,Y,Z]^T$, which can be written as

$$\mathbf{x} = \rho(X\mathbf{p_1} + Y\mathbf{p_2} + \mathbf{p_4}) \tag{48}$$

$$\mathbf{x'} = \mu(X\mathbf{p_1} + Y\mathbf{p_2} + Z\mathbf{p_3} + \mathbf{p_4}) \tag{49}$$

where $\rho$ and are two unknown scalar factors and $\mathbf{p}_i$ is the $i^{th}$ column of $\mathbf{P}$. We set it to be $\mathbf{p_4} = \mathbf{l}/\|\mathbf{l}\|$ for linear independence, and $\mathbf{p_3} = \alpha \mathbf{v_z}$, where $\alpha$ is scale factor (A. Criminisi & I. Reid & A. Zisserman, 1999). Taking the scalar product of Eq. (13) with $\bar{\mathbf{l}}$ yields $\rho = \bar{\mathbf{l}} \cdot \mathbf{x}$, and combining it with Eq. (14), we obtain

$$\alpha Z = -\frac{\|\mathbf{x} \times \mathbf{x'}\|}{(\bar{\mathbf{l}} \cdot \mathbf{x})\|\mathbf{v_z} \times \mathbf{x'}\|} \tag{50}$$

Consequently, if $\alpha$ is known, then a metric value for $Z$ can be easily computed out and conversely, $\alpha$ can be obtained from a reference value $Z_r$. Note that the metric value $Z_r$ defined by $\mathbf{x_r'}$ and $\mathbf{x'}$ also satisfy the Eq. (15).

Thereby, the metric value $Z$ for target can be obtained by the following equation in algebraic representation:

$$Z = \frac{\|\mathbf{x} \times \mathbf{x'}\|}{(\bar{\mathbf{l}} \cdot \mathbf{x})\|\mathbf{v_z} \times \mathbf{x'}\|} Z_r \frac{(\bar{\mathbf{l}} \cdot \mathbf{x_r})\|\mathbf{v_z} \times \mathbf{x_r'}\|}{\|\mathbf{x} \times \mathbf{x_r'}\|} \tag{51}$$

where $Z_r$ is the referred object's metric value.

Herein, the vertical vanishing point $Z_r$ in Eq. (51) is not required in our method which can be obtained from

$$\mathbf{l} = \omega \mathbf{v_z} \tag{52}$$

where $\mathbf{l}$ is the horizon line.

Consequently, the method based on single view metrology only requires two sets of parallel lines to obtain the vanishing line and different combinations of geometric constraints to achieve metric rectification.

**Measurement on vertical and arbitrary plane** In this section, we show that scene measurements on vertical or arbitrary plane can also be retrieved from the camera projection matrix and some scene constraints.

Suppose $\pi_0$ is the reference plane, $\pi_1$ is the vertical plane perpendicular to $\pi_0$ and intersects $\pi_0$ at line $\mathbf{L}$ seeing Figure 2. We denote $\pi_1$ as $\prod_1 = [a_1, b_1, c_1, d_1]^T$, and the point $\mathbf{M}_i$ is in $\pi_1$ if and only if $\pi_1^T \mathbf{M}_i = 0$. Let $\mathbf{l}$ is the corresponding image of $\mathbf{L}$, and $\mathbf{H}$ is the planar homography, then $a_1 = (\mathbf{H^T l})_1$, $b_1 = (\mathbf{H^T l})_2$, $c_1 = 0$, $\pi_1$ ( G. Wang & Z. Hu & F. Wu & H. T. Tsui, 2005). For a point $\mathbf{M}_i$ on the vertical plane $\pi_1$, its coordinates can be retrieved by the intersection of the back-projected ray of image point $\mathbf{m}_i$ and the plane

$\pi_1$, i.e.

$$\begin{cases} s_i \mathbf{m}_i = \mathbf{P} \mathbf{M}_i \\ \Pi_1^T \mathbf{M}_i = 0 \end{cases} \quad (53)$$

where $s_i$ is a scalar factor.

For the measurement on an arbitrary plane, we assume that an arbitrary plane $\pi_2$ intersect the reference plane at line **L** then all the plane passing through **L** can form a pencil which may be expressed as $\Pi_2 = \Pi_1 + \lambda \Pi_0 = [(\mathbf{H^T l})_1, (\mathbf{H^T l})_2, \lambda, (\mathbf{H^T l})_3]^T$, where $\Pi_0 = [0,0,1,0]^T$ is the reference plane and $\Pi_1 = [(\mathbf{H^T l})_1, (\mathbf{H^T l})_2, 0, (\mathbf{H^T l})_3]^T$ is the vertical plane. Therefore, the arbitrary plane is defined up to a unknown parameter $\lambda$ which can be determined from a pair of imaged parallel lines in the plane (G. Wang & Z. Hu & F. Wu & H. T. Tsui, 2005). Once the plane $\pi_2$ is determined, we can take measurements on the plane in a similar way like the vertical plane, using the following equations.

$$\begin{cases} s_j \mathbf{n}_j = \mathbf{P} \mathbf{N}_i \\ \Pi_2^T \mathbf{N}_i = 0 \end{cases} \quad (54)$$

where $s_j$ is a scalar factor, $\mathbf{N}_j$ is a 3D point in $\pi_2$ and $\mathbf{n}_j$ is its corresponding image point.

## FUTURE RESEARCH DIRECTIONS

Image forgeries detection is a comprehensive problem involving computer vision, signal process, machine learning, pattern recognition, to name a few. Future research directions in the field can be concentrated on the two following aspects:

(1) It is necessary to build a consistent model and select the authenticated features for an image to be authenticated, since different features induce diverse authenticating results. The consistent model can be built based on the quality of a natural image, consistency in lighting or imaging, color period and so on.
(2) Object-oriented forensics technology. Images possess not only digital signals but also contexts on human vision and space architecture. Efforts should be made on the changes of an image's context such as appending, removing or modification.

The methods reviewed in this chapter use geometric information to detect tampered regions. In the field of image forgeries detection, traditional methods perform well in some particular circumstances. As a new direction, geometric methods give better detecting results when geometric constraints are available. However, like other algorithms, it also fails when geometric constraints are not at hand. In the future work, it is important to detect the fake regions using less geometric information and make the detecting process automatic.

## CONCLUSION

This chapter presents a new framework for detecting image forgery based on geometrical cues such as cross ratio constraint and shadow geometry constraint. Algorithms have been introduced to obtain different kinds of measurements for fake region detection: 3D height measurement requiring a reference distance; measurements on a vertical or arbitrary plane with respect to the reference plane. The experimental results demonstrate that this method is efficient and can be applied to a variety of target scenes. As a pragmatic and flexible framework, it is also simple and easy to implement. However, it is evident that the problem of detecting digital image forgeries is complicated one with no universally applicable solution. Indeed, it may be easier to undergo forgery processing operations as the editing software improves. The rapid advancements in image editing software make the images undetectable using individual method, and what we need is a large set of methods based on different principles that can be applied to all tampered images. This accumulative evidence may provide a convincing proof that each individual method cannot carry out.

# REFERENCES

Bayram, S., Memon, N., Ramkumar, M., & Sankur, B. (2004). A classifier design for detecting image manipulations. In *Proc. IEEE Int. Conf. on Image Processing, Singapore, 4,* 2645-2648.

Bayram, S., Sencar, H. T. & Memon, N. (2005). Source camera identification based on CFA interpolation. *IEEE International Conference on Image Processing, Genoa, Italy, Sep* (Vol. 3, pp. 69-72).

Cao, X., & Foroosh, H. (2006). Camera Calibration using Symmetric Objects. *IEEE Transactions on Image Processing, 15*(11), 3614-3619.

Chen, M., Fridrich, J., Goljan, M. & Lukas, J. (2008). Determining Image Origin and Integrity Using Sensor Noise. *IEEE Transactions on Information Forensics and Security, 3*(1), 74-90.

Chen, W., Shi, Y., & Su, W. (2007). Image splicing detection using 2D phase congruency and statistical moments of characteristic function. In *Proc. SPIE, Electronic Imaging, Security, Steganography, Watermarking of Multimedia Contents IX, San Jose, CA, J 29 January-1 February, 6505,* 65050R.1-65050R.8.

Cox, I.J., Miller, M.L. & Bloom, J.A. (2002). *Digital Watermarking*. Morgan Kaufmann Publishers.

Criminisi, A., Reid, I., & Zisserman, A. (1999). Single View Metrology. *International Conference on Computer Vision* (pp. 434-441).

Farid, H. & Lyu, S. (2003). *Higher-order wavelet statistics and their application to digital forensics.* IEEE Workshop on Statistical Analysis in Computer Vision.

Farid, H. (2008). *Digital ballistic from jpeg quantization: A follow study* (Technical Report TR2008-638). Department of Computer Science, Dartmouth College.

Farid, H. (2009). A Survey of Image Forgery Detection. *IEEE Signal Processing Magazine, 26*(2), 16-25.

Farid, H. (2009). Seeing is not believing. *IEEE Spectrum, 46*(8), 44-48.

Fridrich, J., Soukal, D., & Lukas, J. (2003, August). *Detection of Copy-Move Forgery in Digital Images.* Pro. Digital Forensic Research Workshop, Cleveland, OH.

Gou, H., Swaminathan, A. & Wu, M. (2007). Noise features for image tampering detection and steganalysis. *IEEE International Conference on Image Processing, San Antonio, TX, 2007* (Vol. 3, pp. 97-100).

Hartley, R. & Zisserman, A. (2004). *Multiple View Geometry in Computer Vision*. Cambridge University Press.

Hsu, Y. F. & Chang, S. F. (2007). Image splicing detection using camera response function consistency and automatic segmentation. *IEEE International Conference on Multimedia and Expo.*

Johnson, M. K. & Farid, H. (2007). Exposing Digital Forgeries in Complex Lighting Environments. *IEEE Transactions on Information Forensics and Security, 2,* 450-461.

Johnson, M. K., & Farid, H. (2007). Detecting Photographic Composites of People. *Proc. IWDW.*

Johnson, M.K., & Farid, H. (2005). *Exposing digital forgeries by detecting inconsistencies in lighting.* ACM Multimedia and Security Workshop, New York, NY.

Johnson, M.K., & Farid, H. (2006). *Metric Measurements on a Plane from a Single Image* (Technical Report, TR2006-579).

Kersten, D., Mamassian, P., & Knill, D. C. (1997). Moving cast shadows induce apparent motion in depth. *Perception, 26,* 171-192.

Liebowitz, D., & A. Zisserman. (1998). Metric Rectification for Perspective Images of Planes. *IEEE Computer Society Conference on Computer Vision and Pattern Recognition.*

Lin, Z., Wang, R., Tang, X., & Shum, H.Y. (2005). Detecting doctored images using camera response normality and consistency. *IEEE Computer Society Conference on Computer Vision and Pattern Recognition* (Vol. 1, pp. 1087- 1092).

Liu, H., Rao, J., & Yao, X. (2008). Feature based watermarking scheme for image authentication. In *Proceedings of the 2007 IEEE International Conference on Multimedia and Expo* (pp. 229-232).

Lukas, J., & Fridrich, J. (2003). *Estimation of primary quantization matrix in double compressed JPEG images.* Digital Forensic Research Workshop.

Lukas, J., Fridrich, J., & Goljan, M. (2005). Determining Digital Image Origin Using Sensor Imperfections. In *Proc. SPIE Electronic Imaging, Image and Video Communication and Processing, San Jose, California., 5685*(2), 249-260.

Lyu & Farid (2005). How Realistic is Photorealistic? *IEEE Transactions on Signal Processing, 53*(2), 845-850.

Mahdian, B. & Saic, S. (2007). Detection of copy-move forgery using a method based on blur moment invariants. *Forensic science international, 171*(2), 180-189.

Ng, T. T. & Chang, S. F. (2004). A model for image splicing. *IEEE International Conference on Image Processing* (pp. 1169-1172).

Ng, T. T., Chang, S. F. & Sun, Q. (2004). Blind detection of photomontage using higher order statistics. In *Proceedings of the 2004 International Symposium on Circuits and Systems, 5,* 688-691.

Pevny, T. & Fridrich, J. (2008). Detection of Double-Compression in JPEG Images for Applications in Steganography. *IEEE Transactions on Information Forensics and Security, 3*(2), 247-258.

Popescu, A. C. & Farid, H. (2005). Exposing digital forgeries by detecting traces of re-sampling. *IEEE Signal Processing Magazine, 53*(2), 758-767.

Popescu, A. C. & Farid, H. (2005). Exposing digital forgeries in color filter array interpolated images. *IEEE Signal Processing Magazine, 53*(10), 3948-3959.

Popescu, A., & Farid, H. (2004). *Exposing digital forgeries by detecting duplicated image regions* (*Technical Report TR2004-515*). Department of Computer Science, Dartmouth College.

Springer, C. E. (1964). *Geometry and Analysis of Projective Spaces*. Freeman.

Swaminathan, A., Wu, M., & Ray Liu, K. J. (2008). Digital Image Forensics via Intrinsic Fingerprints. *IEEE Transactions on Information Forensics and Security, 3*(1), 101-117.

Wang G., Hu, Z., Wu, F. & Tsui, H. T. (2005). Single view metrology from scene constraints. *Image and vision computing, 23*(9), 831-840.

Wang, W. & Farid, H. (2008). *Detecting Re-Projected Video*. International Workshop on Information Hiding.

Zhang, W., Cao, X., Feng, Z., Zhang, J., & Wang, P. (2009). Detecting Photographic Composites Using Two-View Geometrical. *IEEE International Conference on Multimedia and Expo.*

Zhang, W., Cao, X., Zhang, J., Zhu, J., & Wang, P. (2009). Detecting Photographic Composites Using Shadows. *IEEE International Conference on Multimedia and Expo.*

**ADDITIONAL READING SECTION**

Bravo, M.J. & Farid, H. (2008). A Scale Invariant Measure of Image Clutter. *Journal of Vision, 8*(1). 1-9.

Bravo, M.J. & Farid, H. (2009). The Specificity of the Search Template. *Journal of Vision, 9*(1), 1-9.

Bravo, M.J. & Farid, H. (2009). Training Determines the Target Representation for Search. Vision Sciences (VSS), Naples, FL.

Cao, X. & Foroosh, H. (2006). Camera Calibration using Symmetric Objects. *IEEE Transactions on Image Processing, 15*(11), 3614-3619.

Cao, X., & Foroosh, H. (2007). Camera Calibration and Light Source Orientation from Solar Shadows. *Computer Vision and Image Understanding*, 105, 60–72.

Cao, X., Shen, Y., Shah, M., & Foroosh, H. (2005). Single View Compositing with Shadows. *The Visual Computer, 21*(8-10), 639-648. Also in Pacific Graphics 2005.

Cao, X., Wu, L., Xiao, J., Foroosh, H., Zhu, J., & Li, X. (2009).Video Synchronization and Its Application on Object Transfer. *Image and Vision Computing* (In press)

Farid, H. (2008). *Digital Image Forensics*. American Academy of Forensic Sciences. Washington. DC.

Farid, H. (2008). *Photography Changes What We Are Willing To Believe*. Smithsonian Photography Initiative.

Farid, H. (2009). Digital Doctoring: can we trust photographs? In *Deception: Methods, Motives, Contexts and Consequences*.

Farid, H. (2009). Digital Imaging. In *Encyclopedia of Perception*.
Farid, H. (2009). Exposing Digital Forgeries from JPEG Ghosts. *IEEE Transactions on Information Forensics and Security, 4*(1),154-160.
Farid, H. (2009). Photo Fakery and Forensics. *Advances in Computers*, 77.
Farid, H. (2009). Seeing Is Not Believing. *IEEE Spectrum, 46*(8), 44-48.
Farid, H., & Woodward, J.B. (2007). *Video Stabilization and Enhancement* (Technical Report, TR2007-605). Dartmouth College, Computer Science.
Johnson, M. K. (2007). *Lighting and Optical Tools for Image Forensics*. Ph.D. Dissertation, Department of Computer Science, Dartmouth College.
McPeek, M.A., Shen, L., & Farid, H. (2009). The Correlated Evolution of 3-Dimensional Reproductive Structure Between Male and Female Damselflies. *Evolution, 63*(1), 73-83.
Shen, L., Farid, H., & McPeek, M.A. (2009). Modeling 3-Dimensional Morphological Structures using Spherical Harmonics. *Evolution*, *63*(4), 1003-1016.
Wang, W. (2009). *Digital Video Forensics*. Ph.D. Dissertation, Department of Computer Science. Dartmouth College.
Wang, W., & Farid, H. (2009). Exposing Digital Forgeries in Video by Detecting Double Quantization. ACM Multimedia and Security Workshop, Princeton, NJ.

## KEY TERMS AND DEFINITIONS

**Digital forensics**: Authenticate a digital image's integrity.
**Single view metrology**: Get measurements in a single image.
**Planar homology**: A plane projective transformation is a planar homology if it has a line of fixed points together with a fixed point.
**Metric rectification**: Remove the projective distortion from a perspective image.
**Shadow geometry**: The geometric relationship between the shadow point, the light source and the shadow casting object.
**Metric measurement**: Get the measurements in a metric rectified image.
**Region of interest**: The region in an image whose integrity is suspicious.